\documentclass{article}
\usepackage{ijcai26}

\usepackage{xcolor}
\usepackage{cite}
\usepackage{microtype}
\usepackage{graphicx}
\usepackage{url}
\usepackage{hyperref}

\usepackage{booktabs}
\usepackage{multirow}
\usepackage{tabularx}
\usepackage{adjustbox}
\usepackage{array}
\usepackage{tikz}

\hypersetup{
    colorlinks=true,
    linkcolor=blue,
    citecolor=blue,
    urlcolor=cyan,
    pdftitle={Your Paper Title}
}

\title{Acharya~2.0: An AI Tutor for Self-Learning Subjects in Hindi}

\author{
Shantipriya Parida$^{1}$,
Debasish Dhal$^{2}$,
Sonal Khosla$^{2}$, \\
Sambit Sekhar$^{2}$,
Muhammed Abdur Rahmaan Kamaldeen$^{2}$\\
\affiliations
$^{1}$AMD Silo AI, Finland\\
$^{2}$Odia Generative AI, India\\
}

\begin{document}
\maketitle
\begin{center}
\textbf{Abstract}
\end{center}
Text processing is crucial for low-resource languages as it enables the development of tools and technologies that bridge the digital divide and provide equitable access to information and services. By empowering low-resource languages with robust text processing, we can preserve linguistic diversity, enhance education, improve communication, and foster inclusivity in a rapidly digitizing world. Due to the lack of sufficient annotated datasets and linguistic resources, it is challenging to develop accurate and reliable language technologies for these low-resource languages. 

Large Language Models (LLMs) have attracted significant attention with their performance and the value they produce. However, with the ever-evolving language, adapting LLMs for a specific task is time-consuming and resource-intensive. Thus, Retrieval-Augmented Generation (RAG) has overcome this limitation by augmenting LLMs with external knowledge, thereby improving their efficiency and extensive task-specific adaptation. 

Most applications of the RAG technique have been limited to English and has seen limited application in Indic languages. This paper introduces the RAG-powered AI-Tutor application, \textit{Acharya~2.0}, designed to produce academic assistance to Hindi-speaking students. The client-server-based web application integrates many features to support students in accessing and interacting with educational content.     

\section{Introduction}
In a predominantly digital world, text processing for low-resource languages is a critical field of research and application. With insufficient digital resources for low-resource languages, there are few robust AI systems capable of understanding and generating text, creating barriers to education, communication, and technological inclusion for these languages.

In the education domain, text processing can play a transformative role by assisting in the creation of accessible learning materials, real-time translation tools, and intelligent tutoring systems. By using advanced natural language processing (NLP) techniques, text processing can empower students in low-resource language communities with personalized, contextually relevant educational support \cite{soodan2024ai,dan2023educhat}. Intelligent tutoring systems have been shown to provide personalized assistance to students, and enhance learning efficiency. \textit{eTeacher} is one such Intelligent Tutoring System (ITS) that provides personalized assistance to E-learning students. It can automatically identify a student's learning style based on their interactions within an e-learning system through Bayesian networks. Subsequently, eTeacher utilizes the student's profile to offer personalized guidance, suggesting customized courses of action to enhance the learning experience \cite{SCHIAFFINO20081744}.

The paper by Ahuja et al.  discusses in detail the need for ITS and their recent progress \cite{ahuja2013critical}. Additionally, the paper identifies prospective areas for future ITS development, underscoring the significance of human-computer interaction as a promising domain that requires intervention by researchers. DeepTutor\footnote{\url{deeptutor.org}}, a conversational intelligent tutor, is developed as a web application accessible 24/7 through a browser on any device connected to the Internet. 

Developing an LLM-based education system enables personalized student interactions to assess mastery levels and learning preferences. Research suggests LLMs can simulate human behavior, aiding in student modeling and providing data to improve teaching skills for students with limited interaction records. In addition, combining multimodal information provides a general intelligent educational tool \cite{li2023adapting}.

The adaptability of LLMs has driven interest in automating tutoring, particularly interactive mathematical question answering. Though at times, LLM responses can be inaccurate or fail to align with specific educational curricula. Retrieval-Augmented Generation (RAG) has therefore been adopted to improve the relevance and accuracy of generated responses \cite{levonian2023retrieval}.

While several ITSs exist for English, such tools remain scarce for Indic languages. Despite India's rich linguistic diversity, the lack of educational resources in regional languages limits accessibility and inclusivity. Recognizing this need, the Indian government has promoted the use of local languages in education. Motivated by these efforts, this paper presents an AI tutor for the Hindi language that provides interaction in the form of question-answering etc.

\section{Tools and Technologies}

LLMs are powerful AI models capable of understanding and generating human-like text across diverse tasks \cite{ye2023comprehensive}. As training them from scratch is computationally expensive, fine-tuning provides a more efficient alternative by adapting pre-trained models to specific domains with lower cost and faster deployment. Though, LLMs are increasingly being explored for educational applications \cite{extance2023chatgpt}, they often face challenges related to accuracy, relevance, and up-to-date knowledge, which can be addressed using Retrieval-Augmented Generation (RAG) \cite{finardi2024chronicles}.

RAG presents a promising approach to building an AI tutor in education. RAG uses LLMs to understand and generate contextually relevant responses to user queries, incorporating external knowledge sources thereby increasing response quality \cite{levonian2023retrieval}. It also ensures that the educational content remains up to date and comprehensive. Vector embeddings represent text while capturing semantic relationships between words, enabling models to leverage semantic similarity for better generalization, accurate predictions, and effective learning of complex linguistic patterns. 
Additionally, RAG provides access to an external educational knowledge base at inference time. This reduces the need to encode all curriculum knowledge into model weights and makes content updates easier.


In retrieval systems, it is essential to balance “effectiveness” and “efficiency” while considering resource usage (disk/RAM/GPU) \cite{finardi2024chronicles}. RAG Retriever plays a crucial role in enhancing LLM performance by providing relevant context or information during the generation process. It responds to the user queries or prompts and thus has a better understanding of the context.

\section{Architecture}

\begin{figure}[!htb]
\begin{center}
\includegraphics[scale=0.50]{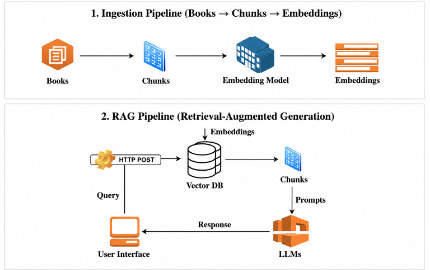} 
\caption{Acharya~2.0 Architecture}
\label{rag_acl_arch}
\end{center}
\end{figure}

The overall architecture of the \textit{Acharya~2.0} web application is shown in Fig. \ref{rag_acl_arch}. The client-server-based architecture consists of several components and is integrated into a Web Application that supports its usage on both desktops and smartphones for ease of use. The major components are the retriever and LLM, upon which the system's performance largely depends.

Some of the key features of \textit{Acharya~2.0} are:

\begin{itemize}
    \item Login through Gmail account.
    \item Supports Hindi subjects from classes 1 to 12 of Central Board of Secondary Education.
    \item User session management which stores user activities.
\end{itemize}

\section{Data Preparation}
The data for the RAG database was scraped from publicly available CBSE (Central Board of Secondary Education) \footnote{\url{https://www.cbse.gov.in/}}, which is the primary board of education in India. The textbooks range from classes 1 to 12, covering a variety of subjects, including Accountancy, Biology, Business Studies, Chemistry, Physics etc., ensuring the complete curriculum is covered. This was achieved by using \textit{pytesseract} (ver. 0.3.10), a Python wrapper of the Tesseract-OCR engine. The extracted data was manually reviewed to remove spelling and punctuation errors that may have been introduced during text extraction. Using OCR, 4.3M tokens spanning 13 distinct subjects (59 subjects across all classes) were added to the vector database.

\begin{figure}[!htb]
\begin{center}

\includegraphics[scale=0.35]{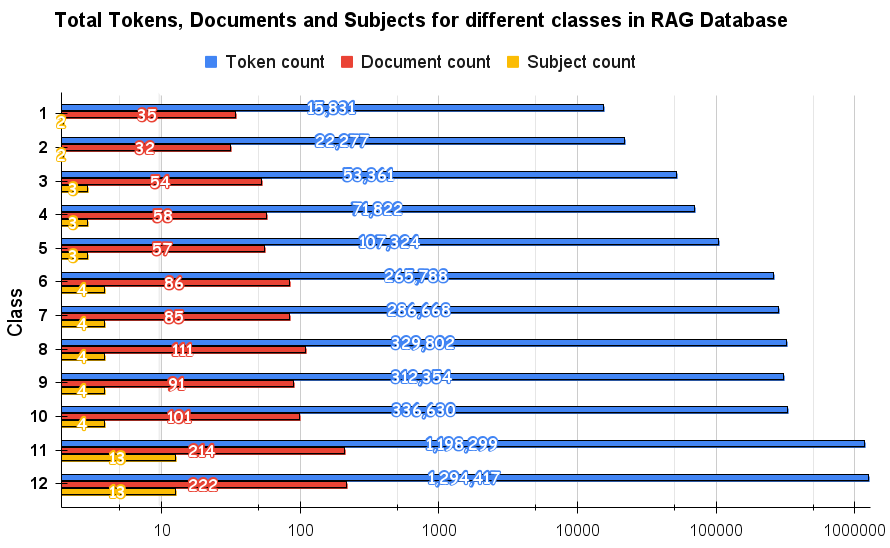}

\caption{\centering RAG Text Database statistics\newline Class 12 has 13 subjects, 222 documents, and 1.29M tokens.}
\label{database-stats}
\end{center}
\end{figure}

\section{Frontend}
We redesigned the interface using React\footnote{\url{https://react.dev/}} and Tailwind CSS\footnote{\url{https://tailwindcss.com/}} to provide a more user-friendly experience. Google login was integrated using react-oauth\footnote{\url{https://www.npmjs.com/package/@react-oauth/google}}, enabling secure authentication through Google Identity Services. We also added Hindi text input using Google Transliteration\footnote{\url{https://developers.google.com/transliterate/v1/getting_started}}, improving accessibility and usability. Additionally, chat history support allows users to revisit previous conversations and their insights.

\section{Backend}

\textbf{Embeddings and Vectorstore} The raw data is converted into embeddings after chunking, to be stored in the vector store for similarity search using the cosine similarity mechanism. We have used sentence transformers (intfloat/multilingual-e5-large-instruct transformer) to generate the embeddings and stored them in ChromaDB's persist directory for subsequent retrieval.\vspace{3mm}

\noindent \textbf{Integration and Operation} The server loads the language and embedding models during startup. User authentication is handled through the '/google/oauth/token' route, which securely manages Google OAuth tokens and user information. After login, users configure their classes and subjects via the '/setting' route, where the corresponding ChromaDB collections are assigned. Queries are processed through the '/home' route: user input is converted into embeddings, the top three relevant documents are retrieved from ChromaDB, and the retrieved context, along with the query, is passed to the language model using a prompt template to generate the response. Finally, the '/logout' route securely signs out the user.

\vspace{3mm}

\noindent \textbf{LLM} For Hindi LLM, we have explored \textit{i)} Mistral-7b and \textit{ii)} Gemma-2B fine-tuned Hindi instruction sets for use in \textit{Acharya~2.0}.
The Mistral-7B model \footnote{\url{https://huggingface.co/OdiaGenAI/mistral_hindi_7b_base_v1}} was fine-tuned using a dataset of 80K Hindi instructions including domain knowledge covering various domains such as: \textit{art}, \textit{sports}, \textit{environment}, \textit{history}, \textit{food}, \textit{music} etc. The Gemma-2B \footnote{\url{https://huggingface.co/blog/gemma}} was fine-tuned using a large instruction set of 188k by combining several instruction sets available in Hindi. The Gemma-2B was selected as the base model based on its performance in comparison with several benchmark tasks \cite{li2024evaluating}. Based on the comparison of both LLMs, we have used the fine-tuned Mistral for \textit{Acharya~2.0} integration.

\vspace{3mm}

\noindent \textbf{API integration} The backend of our system exposes four main FastAPI endpoints. The /loginroute endpoint is used here to handle user authentication via POST requests; when an end-user tries to log in with their credentials, the API tries to verify this, if they are correct then it grants them access. 


\section{Evaluation}
The performance of \textit{Acharya~2.0} has been evaluated using both automated and human evaluation methods covering a diverse array of topics and content. It provides a comprehensive evaluation of Acharya~2.0’s performance. All automated metrics are on a scale of 0 to 1.


     

\subsection{Automated Evaluation}
Three key metrics were used to evaluate the model's outputs presented in Table \ref{tab:automated-scores}. The metrics are \textit{BERT score} \cite{Zhang:2020}, \textit{Cosine Similarity} \footnote{\url{https://scikit-learn.org/stable/modules/generated/sklearn.metrics.pairwise.cosine_similarity.html}}, and \textit{RAGAS score} \footnote{\url{https://docs.ragas.io/en/stable/concepts/metrics/available_metrics/}}. \textit{BERT score} compares each token of the model's output with each token of the ground truth and generates a score. \textit{Cosine Similarity} takes the cosine of the angle between the vector representation of the model's output and ground truth. \textit{RAGAS score} metric is tailor-made for the evaluation of RAG-based models.

RAG Assessment provides a series of metrics, tailor-made for evaluating RAG-derived results. We're using three such metrics: \textit{Context Precision}, \textit{Answer Relevancy} and \textit{Faithfulness}.


\begin{table}[t]
\centering
\renewcommand{\arraystretch}{0.9}
\setlength{\tabcolsep}{3pt}

\begin{tabular}{lcc}
\toprule
\textbf{Metric} & \textbf{Mistral} & \textbf{Gemma} \\
\midrule
\multicolumn{3}{c}{\textit{BERT}} \\
Precision          & 0.706 & 0.681 \\
Recall             & 0.699 & 0.629 \\
F1-Score           & 0.698 & 0.652 \\
Cosine Similarity  & 0.545 & 0.202 \\
\midrule
\multicolumn{3}{c}{\textit{RAGAS}} \\
Faithfulness       & 0.481 & 0.407 \\
Context Prec.      & 0.275 & 0.487 \\
Answer Relevancy   & 0.247 & 0.321 \\
\bottomrule
\end{tabular}

\caption{Automated evaluation scores for Mistral- and Gemma-based RAG models.}
\label{tab:automated-scores}
\end{table}

\subsection{Human Evaluation}
\begin{figure}[!htb]
\begin{center}
\includegraphics[scale=0.37]{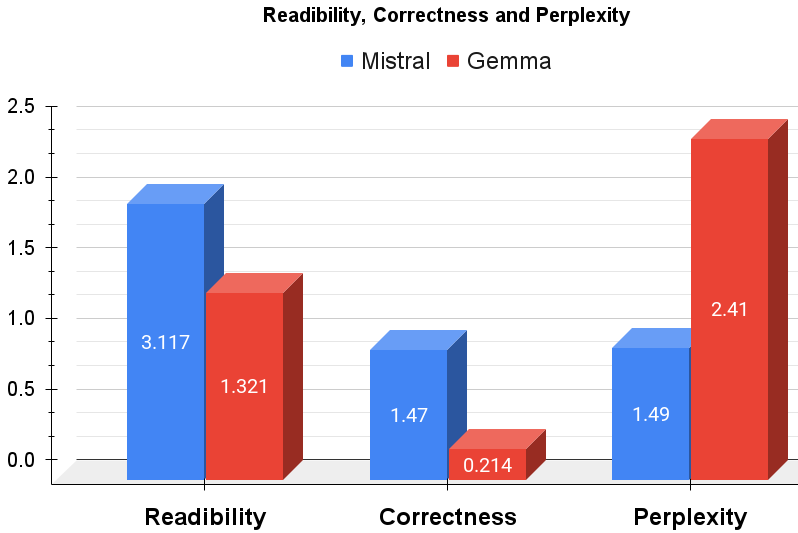} 
\caption{Human Evaluation scores for Mistral and Gemma-based RAG-models}
\label{human-eval}
\end{center}
\end{figure}

We manually evaluated 50 question–answer pairs using three metrics: Readability, Correctness, and Perplexity (0–5 scale). Higher Readability and Correctness scores indicate better performance, whereas a lower Perplexity score indicates fewer hallucinations. The number of annotators were 3, and all of them were native-Hindi speakers who have carried out similar evaluations previously. The scores of all the 3 evaluators have been averaged. The results are shown in Figure \ref{human-eval}.

Mistral performs better than the Gemma model, in most metrics. We believe this is mostly attributed to the difference in size of both models; Mistral is a 7B model, while Gemma is a 2B model. But the Gemma tokenizer is more efficient for Indic languages and thus can embed Indic sentences with fewer tokens than the Mistral tokenizer. Further investigation is needed on this.



\subsection{Strength and Weakness}

\textit{Strengths:}
\begin{itemize}
    \item The Mistral-based RAG model works well for a broad range of questions and answers with reasonable performance.
    \item Both models have minimal inference time.
    \item The Mistral-based model's strengths are in Mathematics and History.
\end{itemize}

\textit{Weaknesses:}
\begin{itemize}
    \item The Gemma-based RAG model produces many erroneous and empty responses, with many characters missing from the beginning and end of words. We believe that improvements in the retriever can rectify it.
    \item Some responses are fully readable and devoid of hallucinations, but are factually incorrect.
\end{itemize}

\section{Analysis}
\textit{Acharya~2.0} works fairly well and achieves good evaluation results as discussed in the previous section. It works primarily on twelve core subjects in the Indian
education system, including Mathematics, Science, Hindi, History, and Geography for Classes 1 to 12. Though both the models have low inference latency, overall the Mistral model performs better than the Gemma model. The best performance has been seen in subjects like Mathematics and History.

\section{Conclusion}

The paper presents \textit{Acharya~2.0}, an LLM-based AI Tutor developed for self-learning subjects in Hindi for high school students, freely available to use. 

The future work includes: \textit{i)} extending \textit{Acharya} to other Indic languages \textit{ii)} improving retriever performance using embedding fine-tuning for RAG \cite{gao2023retrieval}, re-ranking \cite{glass2022re2g}, \textit{iii)} incorporating OCR to embed the images and tables, \textit{iv)} meta-data inclusion, and \textit{v)} supporting a speech-to-text engine for Hindi.  

\section*{Ethics Statement}
\small

We do not envisage any ethical concerns. The dataset used does not contain any personal or personally identifiable information; the source data is publicly available, and there are no risks or harm associated with its usage.

\bibliography{custom}





\clearpage
\appendix

\section{User Interface}
\label{sec:user_interface}

\begin{center}

\includegraphics[width=\linewidth,height=0.32\textheight,keepaspectratio]{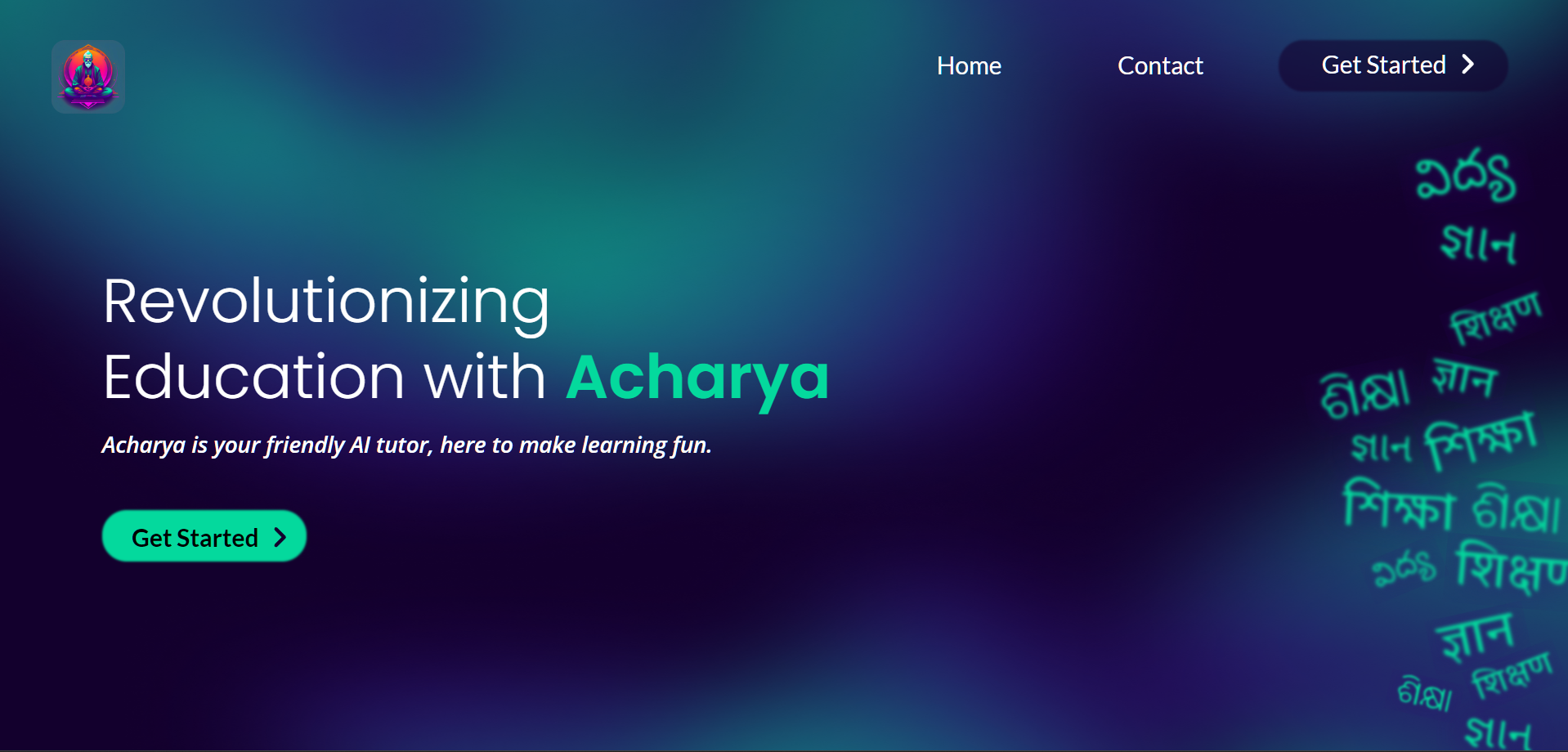}

\vspace{0.1em}

\includegraphics[width=\linewidth,height=0.32\textheight,keepaspectratio]{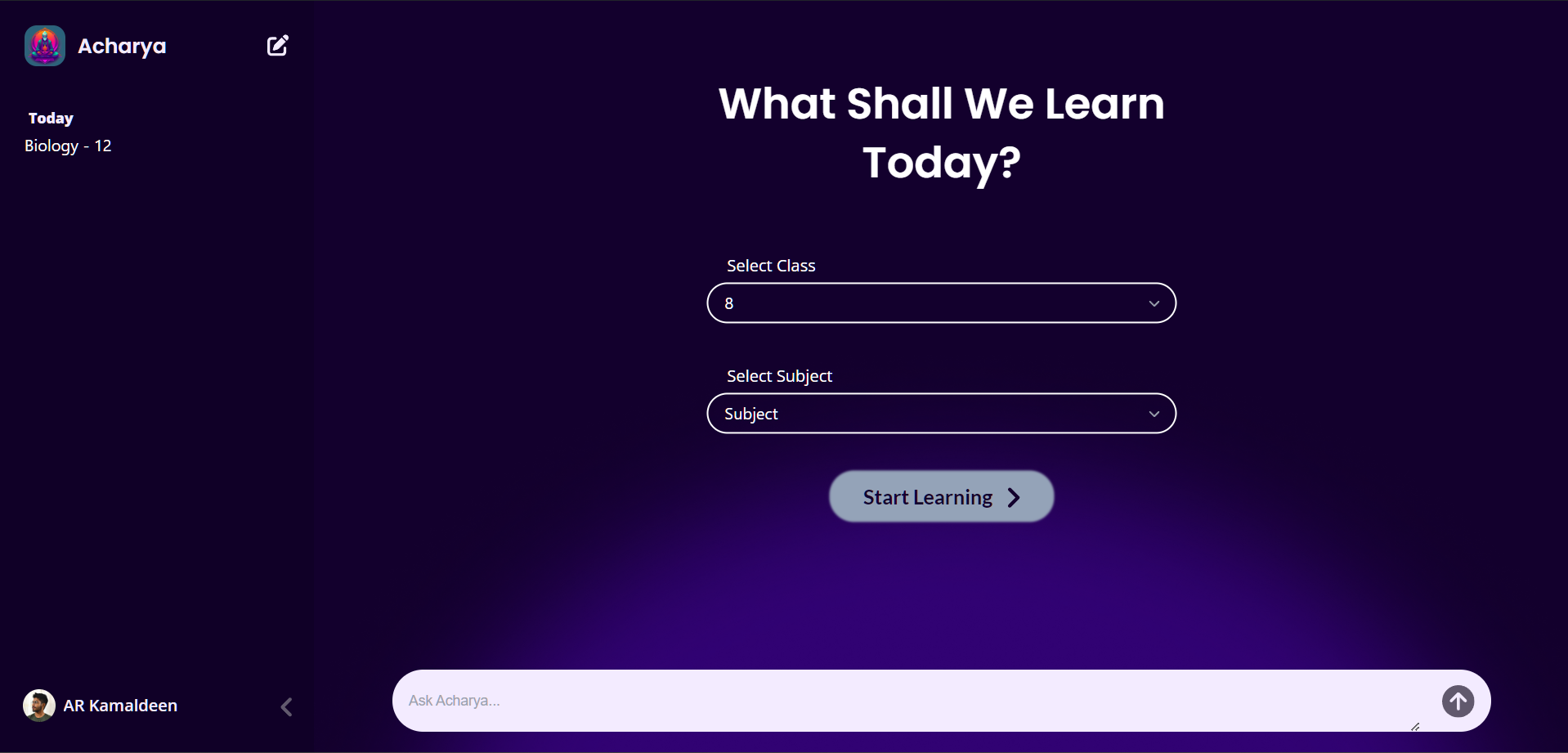}

\vspace{0.1em}

\includegraphics[width=\linewidth,height=0.32\textheight,keepaspectratio]{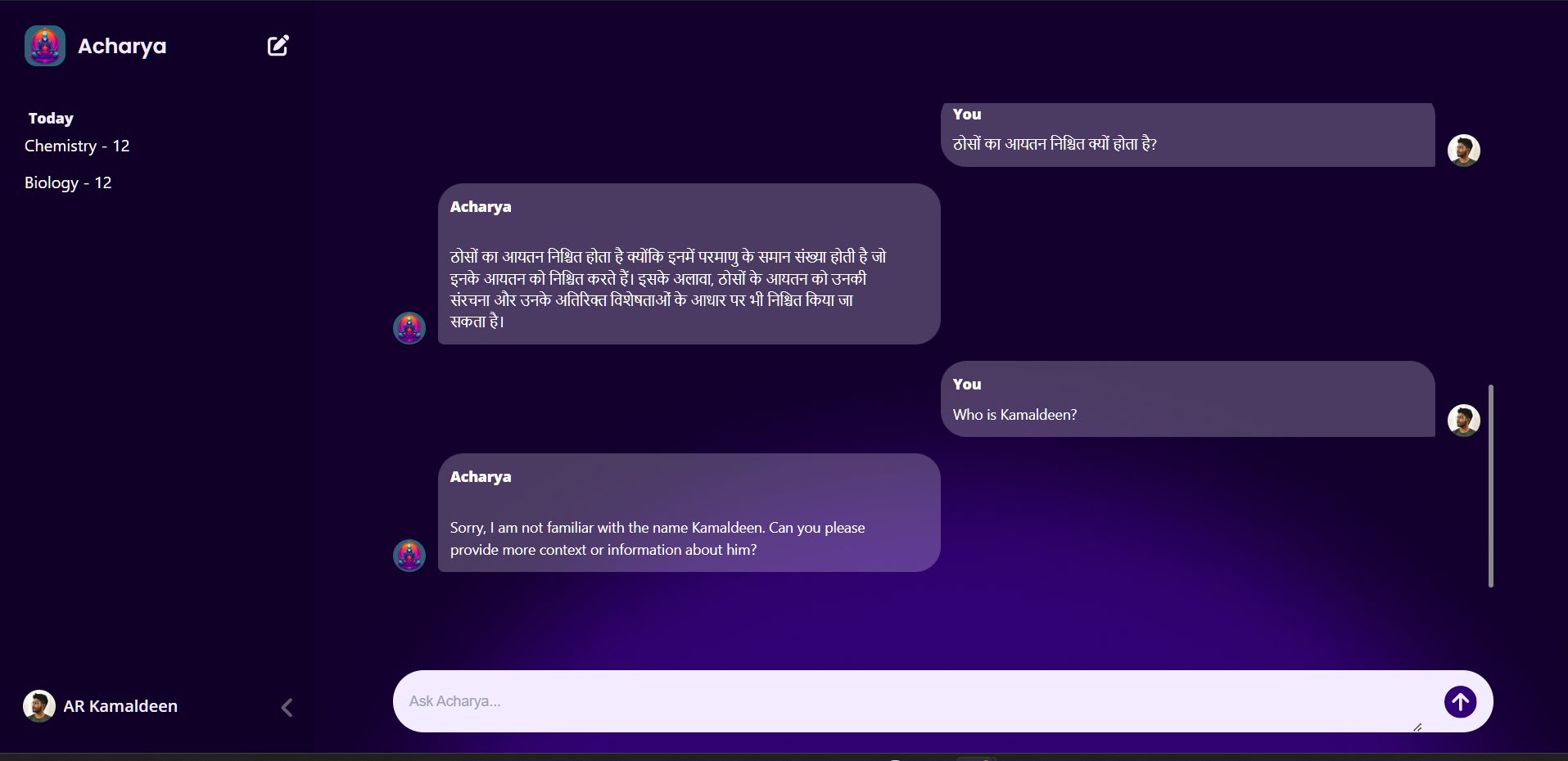}

\vspace{0.1em}

\small
Figure A1: Acharya~2.0 user interface. From top to bottom: (a) landing page, (b) class and subject selection, and (c) generated response.

\end{center}



\end{document}